%% file: tmlr.tex
\documentclass[10pt]{article} 
\usepackage[preprint]{tmlr}

\input{math_commands.tex}

\usepackage{hyperref}
\usepackage{url}
\usepackage{graphicx}
\usepackage[linesnumbered,ruled,vlined]{algorithm2e}
\usepackage{amsmath}
\usepackage{bbm}
\usepackage{multirow}
\usepackage{array}
\usepackage{caption}
\usepackage{subcaption}
\usepackage{makecell} 
\usepackage{enumitem}
\usepackage{placeins}
\usepackage{adjustbox} 

\title{Quantifying Context Bias in Domain Adaptation for Object Detection}

\author{\name Hojun Son \email hojunson@umich.edu \\
      \addr University of Michigan Transportation Research Institute\\
      University of Michigan
      \AND
      \name Asma Almutairi \email asmaalm@umich.edu \\
      \addr University of Michigan Transportation Research Institute\\
      University of Michigan
      \AND
      \name Arpan Kusari \email kusari@umich.edu\\
      \addr University of Michigan Transportation Research Institute\\
      University of Michigan}

\def\month{MM}  
\def\year{YYYY} 
\def\openreview{\url{https://openreview.net/forum?id=XXXX}} 

\begin{document}

\maketitle

\begin{abstract}
Domain adaptation for object detection (DAOD) has become essential to counter performance degradation caused by distribution shifts between training and deployment domains. However, a critical factor influencing DAOD—context bias resulting from learned foreground-background (FG–BG) associations—has remained underexplored. In this work, we present the first comprehensive empirical and causal analysis specifically targeting context bias in DAOD. 
We address three key questions regarding FG-BG associations in object detection: (a) are FG-BG associations encoded during the training, (b) is there a causal relationship between FG-BG associations and detection performance, and (c) is there an effect of FG-BG association on DAOD.  
To examine how models capture FG–BG associations, we analyze class-wise and feature-wise performance degradation using background masking and feature perturbation, measured via change in accuracies (defined as drop rate). To explore the causal role of FG–BG associations, we apply do-calculus on FG–BG pairs guided by class activation mapping (CAM). To quantify the causal influence of FG–BG associations across domains, we propose a novel metric—domain association gradient—defined as the ratio of drop rate to maximum mean discrepancy (MMD).
Through systematic experiments involving background masking, feature-level perturbations, and CAM, we reveal that convolution-based object detection models encode FG–BG associations. These associations substantially impact detection performance, particularly under domain shifts where background information significantly diverges. 
Our results demonstrate that context bias not only exists but causally undermines the generalization capabilities of object detection models across domains. Furthermore, we validate these findings across multiple models and datasets, including state-of-the-art architectures such as ALDI++.
This study highlights the necessity of addressing context bias explicitly in DAOD frameworks, providing insights that pave the way for developing more robust and generalizable object detection systems.
\end{abstract}

\section{Introduction}
\label{sec:intro}

Domain adaptation for object detection (DAOD) has been studied extensively to enable object detectors to perform well on datasets with distribution shifts from the training data \citep{kay2024align, chen2022learning, deng2021unbiased, hoyer2023mic, li2022cross, koh2021wilds, kalluri2023geonet}. It is well known that there's an entanglement between background and foreground features in object detection, leading to a phenomenon called context bias in DAOD \citep{torralba2011unbiased, divvala2009empirical, khosla2012undoing, zhang2024causal, choi2012context, shetty2019not}. Here, significant differences in background features between the source and target domains can cause a notable decline in the quality and number of detections, even when the foreground features remain unchanged. Recent studies in image classification \citep{liadjustment, aniraj2023masking} and segmentation \citep{zhu2024addressing, chen2021scale, dreyer2023revealing} have attempted to mitigate context bias by minimizing this association. \cite{oliva2007role} demonstrated that context bias could result in the corruption of foreground objects by contextually correlated backgrounds, substantially degrading detection quality. However, there has been no prior work specifically analyzing the impact of context bias in DAOD. This work aims to address this gap.

In the realm of human cognition, the brain can accurately and instantly recognize foreground-background (FG-BG) associations without extensive training \citep{papale2018foreground}. Several studies, including \cite{zhang2023different, poort2016texture, papale2018foreground, huang2020source}, have investigated the processes of background suppression and foreground representation to understand the scene and temporal dynamics of foreground and background modulation in the brain. These insights can be applied to the field of computer vision for DAOD through comprehensive analysis of the representation of FG-BG associations.

\begin{figure}[h]
  \centering
   \includegraphics[width=1.0\columnwidth]{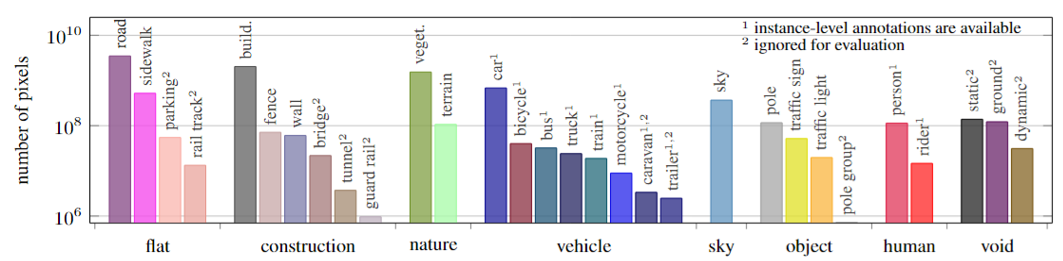}

   \caption{The proportion of background pixels in Cityscapes \cite{cordts2016cityscapes} are the highest of all classes. The image is from Cityscapes publication.}
   \label{fig:road_pixel}
\end{figure}

\begin{figure}[h]
  \centering
   \includegraphics[width=1.\columnwidth]{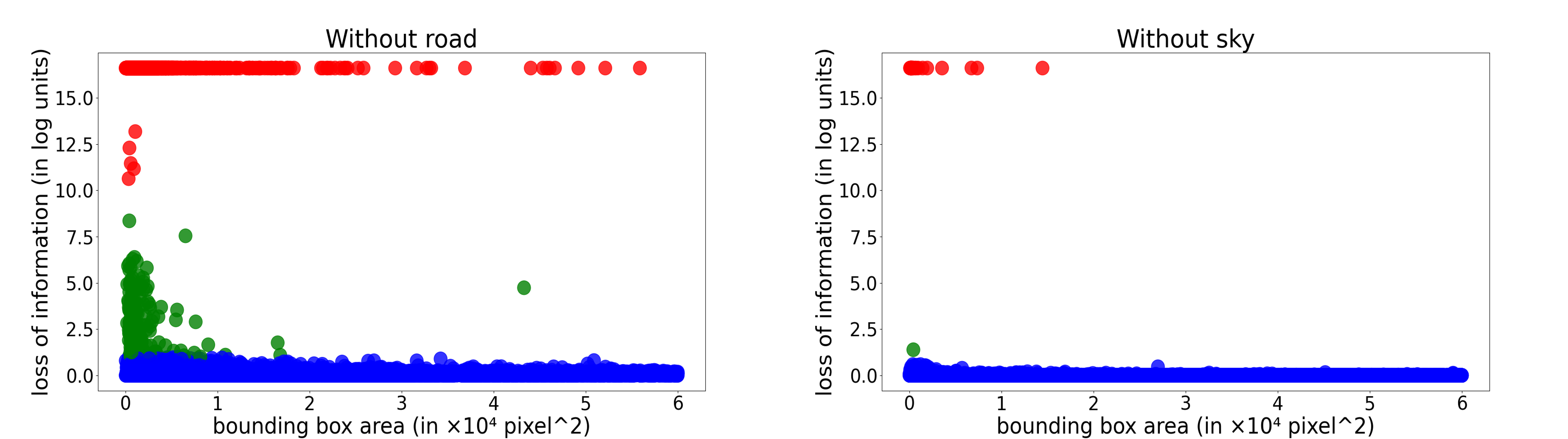}

   \caption{\textbf{Loss of information as a function of the bounding box area of the object.} Left side figure shows the suppression of ``road'' while right side figure shows the suppression of ``sky''. The dots are grouped into three clusters - red indicates missed detections (maximum information loss), green indicates partial matches (significant loss). Blue dots indicate no change. 
   }
   \label{fig:loss_information}
\end{figure}

\begin{figure}[t]
  \centering
   \includegraphics[width=1.0\textwidth]{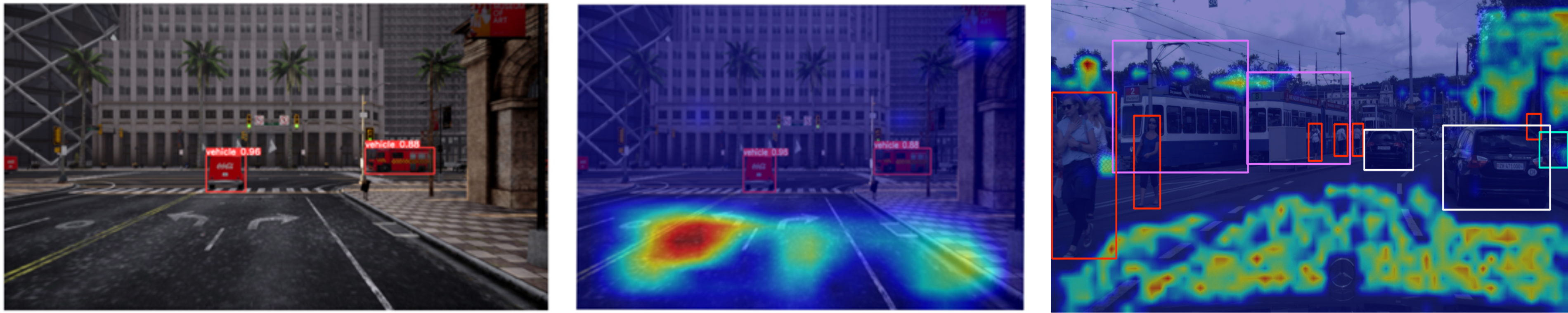}

   \caption{The left figure shows 2D inference of CARLA dataset using YOLOv4 model while the center figure shows the CAM attention map of the inference using EigenCAM \citep{muhammad2020eigen}. The right figure is EigenCAM result of YOLOv11 trained on Cityscapes. Road and vegetation are significantly enhanced.}
   \label{fig:cam_based}
\end{figure}

\subsection{Our observations on context bias}
To motivate our problem, we first looked at the proportion of background features in autonomous driving datasets, as an example. For Cityscapes dataset \citep{cordts2016cityscapes}, the number of pixels from built-up features (such as ``road'' and ``sidewalk'') are much greater than the foreground object pixels (see Fig. \ref{fig:road_pixel}). Based on semantic segmentation outcomes \citep{alonso2021semi, wang2020real, erisen2024sernet}, ``road'' has the highest accuracy and lowest variability.   
As a motivating experiment, we aimed to quantify the change in performance as a function of the background masking for a real dataset. We used the second layer (\texttt{$res2.2$}) of ResNet-50 backbone in the Detectron2 \citep{wu2019detectron2}, trained on the Cityscapes dataset for object detection. We hypothesized that \texttt{$res2.2$} effectively balances low and high-level features on FG-BG. The loss of information was computed when activated features for the specific background regions were zeroed out using semantic labels as a function of the ground truth bounding box area of the foreground objects. We defined performance drop $\Delta IoU$ as $0 \leq \Delta IoU \leq 1$ and the amount of loss of information as the negative log of the complement of $\Delta IoU$ ($-log(\Delta IoU)$). It computed change of intersection over union (IoU) with background masking. Figure \ref{fig:loss_information} shows the performance drop of the removal of ``road'' as opposed to ``sky''. We found that the loss of information is much higher with the ``road'' suppressed as compared to ``sky'', which means that ``road'' has more contextual association with vehicles, particularly when the vehicle size is small.

\begin{figure}[h]
  \centering
   \includegraphics[width=1.\columnwidth, height=0.4\columnwidth]{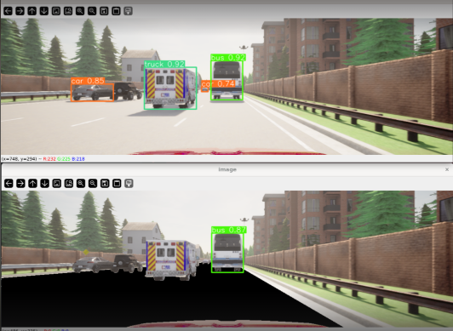}

   \caption{Masking the road on CARLA image and generating inference using YOLOv8 model - top figure shows that 3 out of 4 vehicles get detected correctly while in the bottom figure with road masking, only one vehicle is detected.}
   \label{fig:yolov8 carla}
\end{figure}

\begin{figure}[h]
  \centering
   \includegraphics[width=1.0\textwidth]{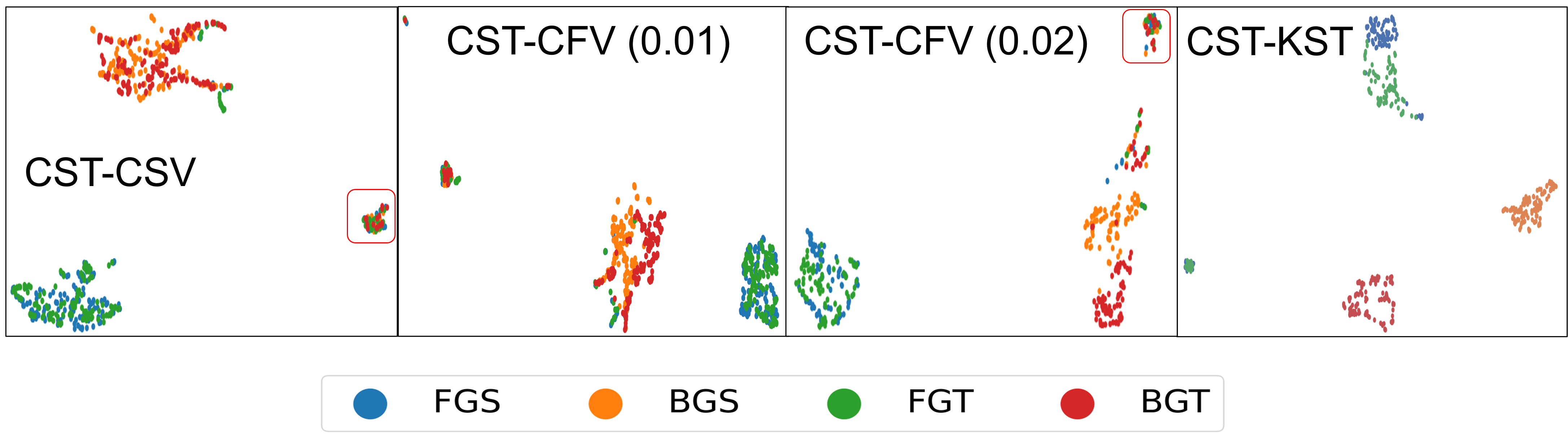}
   \caption{\textbf{UMAP feature embedding results.} ``CST'' is Cityscapes train, ``CSV'' is Cityscapes validation, ``CFV (0.0X)'' are Cityscapes foggy validation with different parameters for fogginess. ``KST'' means KITTI semantic train. We used the ``car'' label, as it is the most common category across the datasets. ``FGS'' is foreground in source domain and ``BGS'' is background in source domain. ``FGT'' and ``BGT'' are foreground and background in target domain.}
\label{fig:2d_umap}
\end{figure}

We trained a YOLOv4 detection model \citep{bochkovskiy2020yolov4} on a sample CARLA \citep{dosovitskiy2017carla} dataset collected under sunny conditions and provided inference on a separate CARLA dataset collected under cloudy conditions. We found that the model was focused on the road in front of the vehicles rather than the vehicles themselves (see Fig. \ref{fig:cam_based}) using class activation mapping (CAM). Additionally, to capture whether the same issue arises across different types of models, we performed an analogous experiment where we transformed the road pixels by masking them and found that YOLOv8 model was unable to detect most of the vehicles otherwise detected in the normal image (see Fig. \ref{fig:yolov8 carla}). This outcome suggests that convolution-based neural network model may have implicitly learned to associate vehicles with road environments, leading to poor performance in detecting vehicles when a different background is present.

In order to understand the spread of feature pattern across domains, we plotted the foreground and background feature distributions of different domains using UMAP \citep{mcinnes2018umap}. Figure \ref{fig:2d_umap} presents the visualization of the foreground and background features from different domains in 2D. We used the features of ``Car'' at the $5^{th}$ ResNet layer (\texttt{$res.5.2$}) from different data distributions. We assumed that \texttt{$res.5.2$} captures high-level features of FG and BG. The interesting finding was the differences of background alignment across the comparisons. It was immediately apparent that as the target domain shifts away from the source domain, the background became more separable than the foreground. For the Cityscapes training (CST) and validation (CSV) dataset, foreground and background features were distinguishable from each other but appeared intermingled between the source and target domains. In the CST-CFV panel, foreground features remained clustered together while the background features were separable but overlapping. We saw an extreme case in the CST-KST where the foreground features between CST and KST were next to each other but were non-overlapping while the background features were very distant from each other. The process to extract features are illustrated in Method section \ref{sec:method}.

Prior studies \citep{choi2012context, torralba2003contextual} have researched context bias for object detection and classification. The studies pointed out that relying only on local features (foreground features in our case) has limitations, including degraded quality due to noise and ambiguity in the target search space. They extended the likelihood to incorporate context information surrounding the foreground, which enhances object classification and detection by providing a stronger conditional probability like the equation \ref{eq:context_priming}. The conditional probability of the object ($O$) given the features ($f$) was given as:
\begin{equation}
    \label{eq:context_priming}
    P(O|f) = P(O|F, B) = \frac{P(F|O,B)P(O|B)}{P(F|B)}
\end{equation}
where $F$ and $B$ are the foreground and background features. However, it did not address DAOD issues like sim-to-real transfer and the root causes of FG-BG associations during training and inference remain unclear, especially given the causal relationships imposed post-detection. 

The challenge with using a convolutional neural network (CNN) assuming identical and independent distribution (i.i.d) to estimate likelihood is the inability to explicitly teach the model to learn each factor in a specific order \citep{scholkopf2021toward, agrawal2019artificial}. In other words, it means that parameters of CNN can be different depending on how it can be trained like the equation \ref{eq:context_priming2}. The modeling can also be interpreted as:

\begin{equation}
    \label{eq:context_priming2}
    \begin{split}
     P(O|f) = P(O|B, F) = \frac{P(B|O,F)P(O|F)}{P(B|F)}
    \end{split}
\end{equation}

In CNNs, likelihood estimation is a process to find the mean of a distribution with proper priors, which necessitates more samples to accurately estimate the true mean. This aligns with the principle that a more extensive and refined dataset, achieved through data augmentation, is crucial for better performance \cite{taylor2018improving}. However, such datasets typically do not account for FG-BG associations, which is subtle to capture during data collection. In summary, FG-BG associations can disrupt the trained estimation process for each probability, leading to performance degradation in target domains due to these broken associations.

From these observations and hypothesis, our fundamental questions are as follows: 
\subsection*{Q1. Are FG-BG associations being inadvertently learned during the training process?}
Deep learning identifies latent patterns that optimize objective functions, typically by maximizing data likelihood. During the feature extraction process, models may learn spurious or unexpected features if such features improve predictive performance without any understanding of causality by following the previous reasoning \citep{bishop2006pattern, goodfellow2016deep, mackay2003information, murphy2012machine}. This underscores the importance of incorporating causal reasoning into deep learning frameworks to improve robustness and generalization \citep{scholkopf2021toward}. From our motivation (see Fig. \ref{fig:loss_information}, \ref{fig:cam_based}, and \ref{fig:yolov8 carla}), we conducted two experiments for Q1 by performing class-wise and feature-wise background removal experiments.

These experiments was designed to capture the existence of FG-BG associations under fair conditions. From these findings about FG-BG associations, we pose the following question: \textbf{To what extent does FG–BG associations affect model accuracy?} 
To address this, we conducted a series of experiments aimed at capturing, representing, and quantifying the impact of FG–BG associations across domains, leading into the next set of questions (Q2 and Q3).

\subsection*{Q2. Is there a causal relationship between FG-BG associations  and object detection?}
\begin{figure}[h]
\begin{center}
\centering
\includegraphics[width=0.9\textwidth]{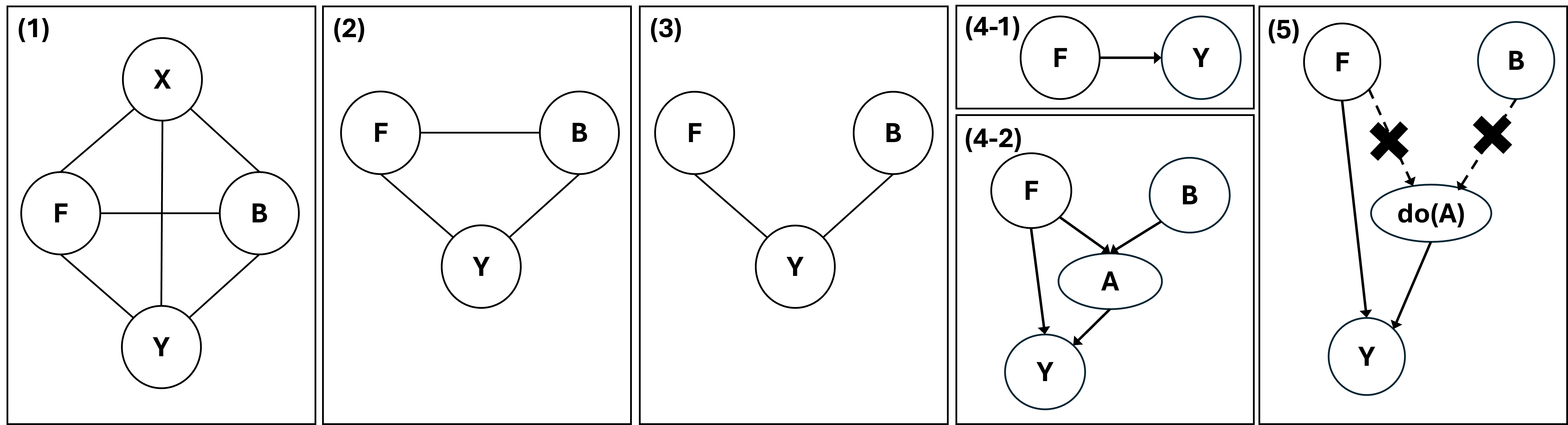}
\end{center}
\caption{\textbf{Identification process to generate causal graph.} (1) is the first step from complete undirected graph where ``X'' is a image, ``F'' is foreground, ``B'' is background, and ``Y'' is a outcome. ``X'' is removed because object detection model will engineer on the input. so it generates (2). (3) is induced because ``F'' and ``B'' are independent. (4-1) is the optimal factor graph we derive that the outcome is solely related to ``F''. (4-2) indicates we captured from preliminary experiments (see Introduction \ref{sec:intro}). ``A'' indicates FG-BG associations. (5) is the factor graph with do-calculus to quantify causal effect of ``A'' on ``Y''. By applying backdoor adjustment, we can compute quantify the causal effect. }
\label{fig:causal graph identification}
\end{figure}
We employed graph-based causal analysis to investigate the causal effect of FG–BG associations on object detection performance. As illustrated in Figure \ref{fig:causal graph identification}, a causal model was constructed using the PC algorithm \cite{glymour2019review} to infer causal relationships within the object detection pipeline. In the context of a graphical casual model (F \textrightarrow Y \textleftarrow B), it represents the joint distribution P(Y, F, B), which can be decomposed as either $P(Y|F,B)P(F|B)P(B)$ or $P(Y|F,B)P(B|F)P(F)$ (see the panel (3) in Fig. \ref{fig:causal graph identification} ,  equations \ref{eq:context_priming} and \ref{eq:context_priming2}). This is a spurious factor leading to FG-BG associations in CNN training. Causal identification further enables through intervention analysis (do-calculus). The causal effect of FG–BG associations ($A$) on object detection performance ($Y$) can be expressed as
$P(Y|\text{do}(A)) = \sum_{F} P(Y|A,F), P(F)$,
following the backdoor adjustment formula. This effect can be quantified by the expected difference
$\mathbb{E}[Y|A=0,F] - \mathbb{E}[Y|A=1,F]$,
which captures the impact of FG–BG association ($A$) on detection outcomes ($Y$), conditioned on contextual features ($F$) (see Fig. \ref{fig:causal graph identification}). To this end, we designed an experiment through intervention via CAM and instance masks. By combining CAM with ground-truth instance masks, we performed do-calculus interventions incurring backdoor adjustment to model the causal influence of FG-BG associations on object detection accuracy. The combination of CAM and instance masks controlled activated background regions for each instance with different threshold.

\subsection*{Q3. What is the impact of FG-BG associations on DAOD and how to quantify the effect?}
While the causal association with detection outcomes ($Y$) learned in source domains typically remains stable, the conditional distribution $P(Y|A)$ may shift due to background distribution changes. This can weaken the causal strength in target domains, resulting in performance degradation. We quantified this effect using the domain association gradient (which we refer to as $Gradient$ in the rest of the paper), which captures the impact of FG–BG associations and using summation of intertwined features within and across domains respectively. For intra-domain analysis, we applied the maximum mean discrepancy (MMD) metric to compare contextual features of instances clustered by associated and non-associated background features. For cross-domain evaluation, we measured the feature discrepancies using MMD of associated and non-associated groups across domains. We would like to note that we termed it as $Gradient$ since it quantifies the change of the response due to the change in FG-BG association.

Our experimental process is described in detail in Section \ref{sec:method}.

\subsection{Contributions}
Our main contributions are as follows:
\begin{itemize}
\item We highlight a crucial gap, suggesting that considering alleviation of context bias is essential for enhancing the generalization and robustness of models across various environments by quantifying its effect. None of the current approaches investigate how context bias can manifest across various domains. We examine the issue of DAOD in relation to context bias. 

\item We analyze FG-BG associations and causal relationship through the drop rate and do-calculus. Additionally, we employ distance-based metrics to measure the association between foreground and background under domain shifts. We also propose an additional metric, domain association gradient, to quantify the context bias on the source and target domain respectively.

\item We provide a novel and practical research perspective by framing context bias as a critical factor in cross-domain object detection. Our study follows a logical progression from empirical observation to theoretical analysis, followed by quantitative and qualitative evaluation resulting in convincing evidence that FG-BG associations significantly affect domain adaptation performance.
\end{itemize}

\section{Related Work}
\label{sec:related}
\subsection{FG-BG Associations and Context Bias}
There has been a number of studies aimed at improving performance in tasks such as classification, object recognition, and object localization researching background influence \citep{xiao2020noise, liang2023impact,zhang2007local, ribeiro2016should, zhu2016object, rosenfeld2018elephant, barbu2019objectnet, sagawa2019distributionally} and context bias \citep{torralba2011unbiased, khosla2012undoing, choi2012context, shetty2019not}.
\cite{xiao2020noise} and  \cite{zhu2016object} studied background effect on accuracy of classification by modifying images with different combinations of foreground and background. \cite{choi2012context} proposed a graphical model which modeled FG-BG associations using conditional probability which serves as a methodological inspiration for us. Several studies have addressed context bias using techniques, such as data augmentation to generate out-of-distributions samples into the background, combination of naturally unmatched background and foreground (e.g., an elephant in room), and applying background removal during training. \cite{torralba2003contextual} demonstrated that background effect can be factorized into object priming, focus of attention, and scale selection by modeling the FG-BG associations in a probabilistic model. 
\cite{liang2023impact} studied background influence using fashion dataset \citep{jia2020fashionpedia, takagi2017makes}. 
These studies \citep{zhai2024background, wu2022background} localized foreground objects better than CAM-based algorithms without using bounding box information and with only classification labels.
These prior works focused on context bias in the same domain and use datasets with smaller variations such as centered objects or single objects. Thus, we found that there remains a gap in understanding how context bias affects DAOD.



\subsection{Domain Adaptation for Object Detection}
Different variations of DAOD methods have been proposed using feature alignment, synthetic images, and self-training or self-distillation. Feature alignment finds transformations between source and target domain to reduce distribution shift with adversarial training \citep{he2019multi, chen2021scale, ganin2016domain, zhu2019adapting}. It can be helpful to extract common latent features from different domains. Progressive Domain Adaptation for Object Detection \citep{hsu2020progressive} synthesized new dataset by using cycleGAN \citep{zhu2017unpaired} which enables to bridge domain gaps and Self-Adversarial Disentangling for Specific Domain Adaptation \citep{zhou2023self} achieved 45.2 mAP on Cityscapes to Cityscapes foggy dataset using synthetic images. \cite{gong2022improving} utilized transformers to focus on aligning features across backbone and decoder networks. However, combining multiple sources into a single dataset and performing single-source domain adaptation for feature alignment does not guarantee better performance compared to using the best individual source domain \citep{zhao2020multi}.

Self-training uses a teacher model to predict pseudo labels on target domains to gradually understand domain shift \citep{caron2021emerging, pham2022revisiting, cai2019exploring, chen2022learning, cao2023contrastive}. MIC \citep{hoyer2023mic} employed masked images on teacher-student model and MRT \citep{zhao2023masked} suggested modified masked based retraining approach on the teacher-student model. \cite{kay2024align} performed an alignment and distillation to enforce invariance across domains to reduce discrepancy of features.

Finding common features from multiple domains is critical for DAOD. They have summarily demonstrated that the foreground features in latent space can be aligned using dimension reduction methods such as UMAP \citep{mcinnes2018umap} and t-SNE \citep{van2008visualizing}. These studies did not address how to manage context bias when adapting across different domains. Instead, they proposed and validated their methods within DAOD framework using accuracy metrics. Thus, we focused on analyzing the root causes of domain discrepancy in object detection both qualitatively and quantitatively.

\section{Method}
\label{sec:method}

The following abbreviations are used throughout this paper to refer to the datasets and models in Table \ref{tab:abb}:
\begin{table}[!h]
    \centering
    \caption{Dataset and model abbreviations}
    \begin{tabular}{c c}
    \hline
    Abbreviation & Meaning \\
    \hline
      CST &  Cityscapes Train\\
      CSV/ CFV / CRV   &  Cityscapes Validation / Foggy / Rainy \\
      KST  &  KITTI Semantic Train \\
      BG-20K  &   Background 20K Dataset \\
        VKC / VKF / VKM /   &   Virtual KITTI Clone / Fog / Morning / \\
       VKO / VKR / VKS & Virtual KITTI Overcast / Rain / Sunset \\
       \hline
      ALDI++ & ResNet-50 FPN with ALDI++ best \\
      Res  &  ResNet-50 FPN  \\
      Eff  &  EfficientNet-B0 FPN \\
      
      \hline
 Yo&YOLOv11
    \end{tabular}
    \label{tab:abb}
\end{table}

\subsection{Models}
We employed ResNet-50 (``Res'') and EfficientNet-B0 (``Eff'') as backbones for FPN models implemented in Detectron2, as well as YOLOv11 (``Yo'') \citep{khanam2024YOLOv11}, an anchor-free detection model. ``Res'' represents a backbone dominantly used in different architectures. ``Eff'' was chosen for its lightweight architecture. Additionally, we included the state-of-the-art DAOD method ALDI++ with a ResNet-50 backbone to evaluate its effectiveness in mitigating FG–BG associations.

\subsection{Datasets}
We used multiple datasets for training and evaluation, including Cityscapes, KITTI Semantic, and various subsets of Virtual KITTI. Additionally, BG-20K, a collection of 20,000 images containing non-salient objects, was utilized to generate randomized background images. The Cityscapes and ``KST'' sets share 8 foreground and 11 background object categories. The Virtual KITTI subsets contain 3 foreground and 10 background object classes.

The dataset sizes are as follows:
\begin{itemize}[topsep=1pt,itemsep=5pt,parsep=0pt,partopsep=0pt]
    \item \textbf{Cityscapes}: 2,950 training images, 500 validation images, 1,500 foggy validation images, and 1,188 rainy validation images.
    \item \textbf{KITTI Semantic Train}: 200 images.
    \item \textbf{Virtual KITTI Semantic}: 2,126 images across 6 simulated weather conditions. This dataset is synthetic and based on object tracking in diverse environments.
\end{itemize}
\subsection{Training and Tests}
We trained ``Res'' and ``Eff'' on the ``CST'', ``KST'' , and ``VKC'' datasets. ``Yo'' was trained with the same condition using Ultralytics frameworks. ALDI++ was trained on (``KST'',``CSV'') and (``VKC'',``VKF'') as source and target domain pairs respectively. We used the pre-trained ALDI++ model provided by the official repository without additional training for Cityscapes. For training, we used a learning rate of 0.02 for ``Res'' with an input resolution of 1024×2048 for Cityscapes and 375×1242 for KITTI-related datasets. For ``Eff'', we used 1024×1024 resolution for Cityscapes and the same KITTI resolution, with a learning rate of 0.01. All models used identical data augmentation: resizing and cropping, color jitter, and horizontal flipping. Each model was trained with a batch size of 8. Training ran for approximately 100 epochs for ALDI++, ``Res'' and ``Yo'', and 200 epochs for ``Eff''. 
During evaluation, we used 1024×2048 images for Cityscapes-related datasets and 375×1242 for KITTI-related datasets on models except ``Yo''. ``Yo'' was trained and evaluated on 512x1024 and 320x1024 image resolution respectively. The best model checkpoint was selected based on the highest mean Average Precision with 0.5 IoU threshold (mAP@50) with respect to DAOD perspective. For example, ``Res'' achieved a mAP@50 of.67.758 on the Cityscapes validation set and 54.617 on the ``CFV'' at epoch 7799, and 67.597 and 57.131 at epoch 7999, respectively. We selected the model from epoch 7999 for subsequent experiments. For ``KST'', the model with the highest performance on ``CFV'' was chosen. For ``VKC'', we selected the model with the highest mAP@50 on ``VKF'', which represents the largest domain shift among the Virtual KITTI variants. All models except ``Yo'' were trained using the standard loss functions provided by Detectron2 and Ultralytics was utilized for ``Yo''. Training was conducted using an NVIDIA RTX A4500 GPU.
\begin{figure}[h]
\begin{center}
\centering
\includegraphics[width=0.95\textwidth]{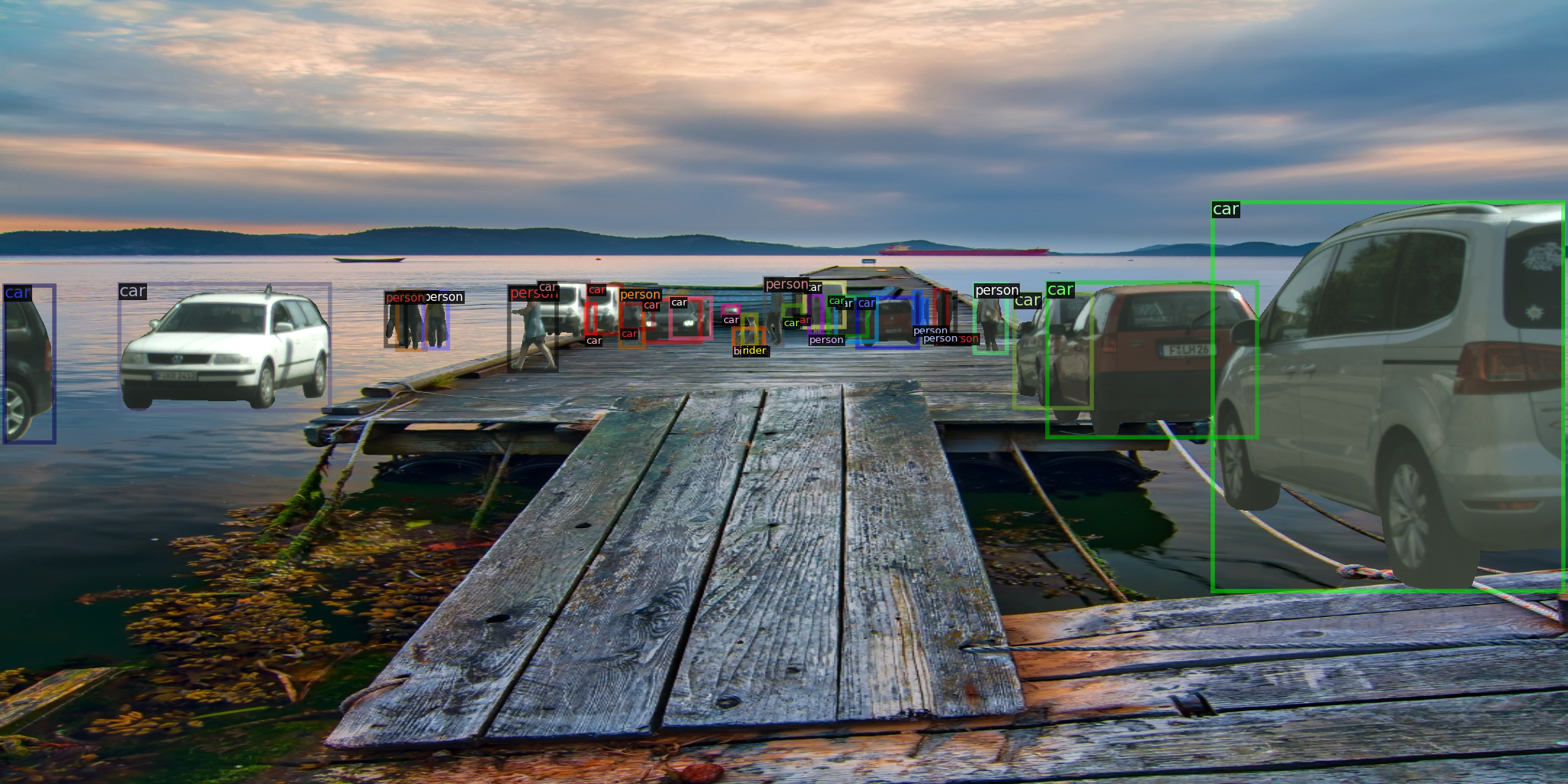}
\end{center}
\caption{\textbf{Example of synthetic image.} We visualize the annotated bounding boxes for foreground objects.}
\label{fig:bg_example}
\end{figure}
\subsection{Q1 - Exp1. Class-wise Background Removal Experiments in Image Space}
The initial experiment evaluates the effect of background variation by performing inference on foreground objects placed over random background images fixed across different domains. Foreground regions were preserved and composited with backgrounds randomly sampled from the BG-20K dataset. The same sequence of background images was applied consistently across domains, thereby reducing the learned association between foreground and background. To ensure statistical validity, the experiment was repeated 6 times using different random sequences of background selections. Algorithm \ref{algo:fg-bg background image space} describes the experiment process. The results, summarized using the mAP@50 metric and standard deviation, are presented in Table \ref{tab:bg_image_space_evaluation}. Figure \ref{fig:bg_example} shows an example of superimposing the foreground objects with a random background image from the BG-20K dataset.

\begin{center}
\begin{minipage}{0.9\linewidth}
\begin{algorithm}[H]
\caption{Class-wise Background Removal Experiments in Image Space}
\label{algo:fg-bg background image space}
\KwIn{
    \begin{itemize}
        \item $FG$: Set of foreground object instances
        \item $BG-{20K}$: Set of 20,000 random background images
        \item $D$: Set of target domains for inference
    \end{itemize}
}
\KwOut{Mean and standard deviation of mAP@50 across 6 repeated trials}
\ForEach{domain $d \in D$}{
    \ForEach{foreground object $f \in FG$}{
          Randomly sample a fixed set of background images $BG_i \subset BG_{20K}$\;

        Synthesize image $I_{f,d}$ by placing $f$ onto a background from $BG_i$\;
    }

    Apply the trained model to all synthesized images $\{I_{f,d}\}$ for domain $d$\;

    Measure detection performance using mAP@50\;
}
Compute mean and standard deviation of mAP@50 across all 6 trials\;
\end{algorithm}
\end{minipage}
\end{center}
\subsection{Q1-Exp2. Feature-wise Background Removal Experiments in Feature Space}
\label{exp2:FG-BG drop rate feature space}
The second experiment investigates feature-wise FG–BG associations by selectively suppressing specific background labels in the feature space during inference. Using ground-truth semantic annotations, a particular background class (e.g., ``road'') was removed in repeated inference runs. This was achieved by zeroing out activation values in corresponding background regions at shallow network layers: \texttt{$res2.2$} in ``Res'' and ALDI++, \texttt{$backbone.bottom\_up.\_blocks.0$} in ``Eff'', and \texttt{$model.1$} in ``Yo''. Due to the hierarchical nature of deep learning models, this targeted suppression weakened the FG–BG associations for the removed background label, potentially affecting detection outcomes. To evaluate this effect, we measured the number of detections from unmodified models and compared them to detections after modification. The drop rate, calculated for each FG–BG class pair, reflects the sensitivity of foreground object detection to the presence of specific background labels.

\textbf{Definition of detection drop:}
A detection was counted as dropped under either of the following conditions:
\begin{enumerate}[topsep=1pt,itemsep=5pt,parsep=0pt,partopsep=0pt]
    \item The predicted class changed due to background label removal.
    \item The loss of information we defined was greater than 1.0, which means IoU with ground truth significantly decrease (less than 0.1 IoU).
    \item The prediction matched a different ground-truth object not originally considered a true positive.
\end{enumerate}

Importantly, drop rates were computed only for true positive cases. To ensure statistical rigor, the process was repeated 6 times. Since drop rate distributions did not satisfy normality assumptions, we employed the Wilcoxon signed-rank test \citep{woolson2005wilcoxon} to assess statistical significance. Algorithm \ref{algo:background removal feature space} describes the experiment process. 

\begin{center}
\begin{minipage}{0.9\linewidth}
\begin{algorithm}[H]
\caption{Feature-wise Background Removal Experiments in Feature Space}
\label{algo:background removal feature space}
\KwIn{
    \begin{itemize}
        \item $FG$: Set of foreground object instances
        \item $BG$: Set of background regions with semantic labels
        \item $L_{remove}$: Background label to be removed (e.g., ``road'')
        \item $M$: Deep learning models (e.g., ``Res'')
    \end{itemize}
}
\KwOut{Drop rate statistics for each FG-BG pair and Wilcoxon test results}

\For{$i = 1$ \KwTo $6$}{
    \ForEach{model $m \in M$}{
        \ForEach{image $x$ with semantic ground truth}{
            Perform standard inference on $x$ with model $m$, store number of detections $D_{std}$\;

            Remove BG activated pixels in shallow feature maps corresponding to label $L_{remove}$ (e.g., \texttt{$res2.2$} for ``Res'')\;

            Perform modified inference on $x$, store number of detections $D_{mod}$\;

            Compute detection drop $\Delta D = D_{std} - D_{mod}$ for each FG-BG pair\;
        }
    }
    Store all $\Delta D$ values for statistical analysis\;
}

Aggregate drop rates for each FG-BG pair across all trials\;

Conduct Wilcoxon signed-rank test to assess significance of drop rate distributions\;
\end{algorithm}
\end{minipage}
\end{center}

\subsubsection{Q2-Exp1. FG-BG Association with Respect to Activated Background Region}
\begin{figure}[t!]
\begin{center}
\centering
\includegraphics[width=0.95\textwidth]{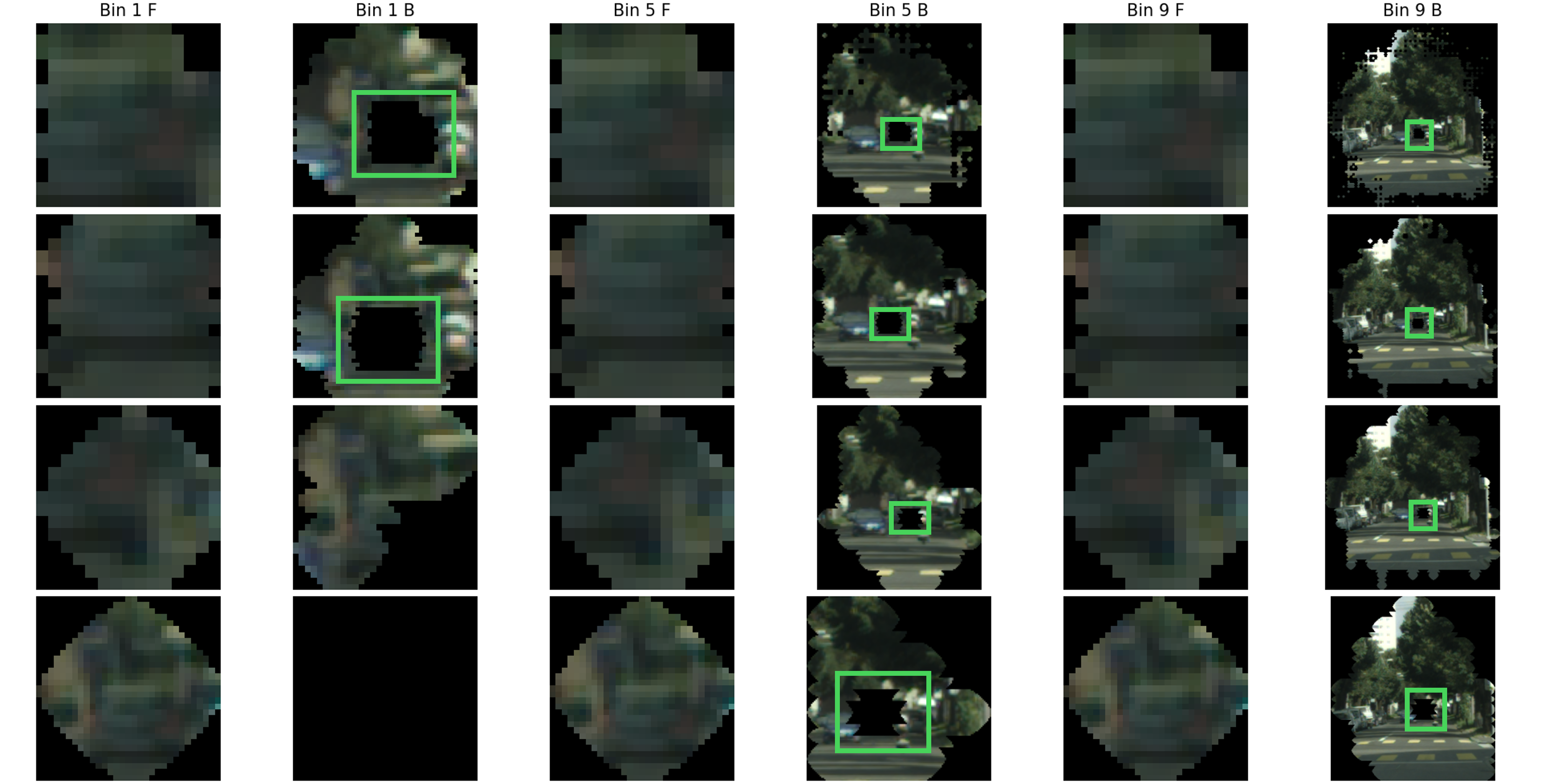}
\end{center}
\caption{\textbf{``Res'' foreground and background images using CAM and instance masks.} Each column corresponds to bin 1, 5, and 9 with foreground and background orders. Each row is different layers. The green bounding boxes highlight where the foreground object gets located. Foreground regions are maintained across layers. The blurriness of the images is due to them being scaled up.}
\label{fig:resnet50_bins}
\end{figure}

Using CAM masks with varying thresholds, we measured the mAP@50 drop rate to investigate the causal influence of activated background regions on object detection performance. Smooth-GradCAM++ \citep{omeiza2019smooth} generates contextually meaningful instance masks by gradient backward on each object's score with 0.85 confidence threshold. The CAM masks were binarized by applying threshold values that decreased by 0.1 with each bin increase. The extent of the activated background region was controlled by the chosen threshold, while the masked foreground region remains fixed throughout the experiment, regardless of background variation. Through statistical analysis, the causality of association and accuracy were demonstrated. Algorithm \ref{algo:CAM FG-BG association} describes the experiment process. Figure \ref{fig:resnet50_bins} illustrates the contextual masks depending on different layers and bins. We define the hit ratio as the ratio of foreground and background pixels captured by CAM in the activation maps, normalized by the number of ground truth pixels from the instance masks. ``FG mean'' is the number of pixels hit by CAM and ground truth of instance mask. ``BG mean'' is activated background region over the number of foreground pixels. We averaged all instances hit ratio to compute ``FG mean'' and ``BG mean'' respectively. It indicates CAM captures properly contextual information for each instance. Definition of associated and non-associated is in Section \ref{sec:ass no-ass features}. 

\begin{center}
\begin{minipage}{0.9\linewidth}
\begin{algorithm}[H]
\caption{Causality Analysis via Smooth-GradCAM++ Mask Thresholding}
\label{algo:CAM FG-BG association}
\KwIn{
    \begin{itemize}
        \item $FG$: Set of foreground object instances with 0.85 confidence threshold
        \item $x$: Input image
        \item $T$: Set of CAM thresholds (e.g., maximum of activation value to $1e^{-9}$ in 0.1 decrements)
    \end{itemize}
}
\KwOut{Drop rates under different CAM thresholds}

\ForEach{foreground instance $f \in FG$}{
    Generate Smooth-GradCAM++ map $H_f$ using prediction confidence of $f$ from image $x$\;

    \ForEach{threshold $t \in T$}{
        Binarized CAM mask: $M_f^{(t)} = \mathbbm{1}(H_f \geq t)$\;

        \textit{(Note: foreground region remains fixed; only background area changes)}\;
        
         Remove partial BG activations in shallow feature maps corresponding to the Masks (e.g., \texttt{$res2.2$} for ``Res'') and compute detection result $D_{mod}$\;

        Compute drop rate: $\Delta D_t = D_{std} - D_{mod}$ where $D_{std}$ is detection without CAM masking\;
    }
}
Aggregate drop rates and hit ratios across all instances\;

Analyze drop rate trend across thresholds to infer FG-BG causality\;

\end{algorithm}
\end{minipage}
\end{center}

\subsection{Q3-Exp1 and Exp2. Quantification the Causal Effect on DAOD}
\begin{figure}[th!]
\begin{center}
\centering
\includegraphics[width=0.9\textwidth]{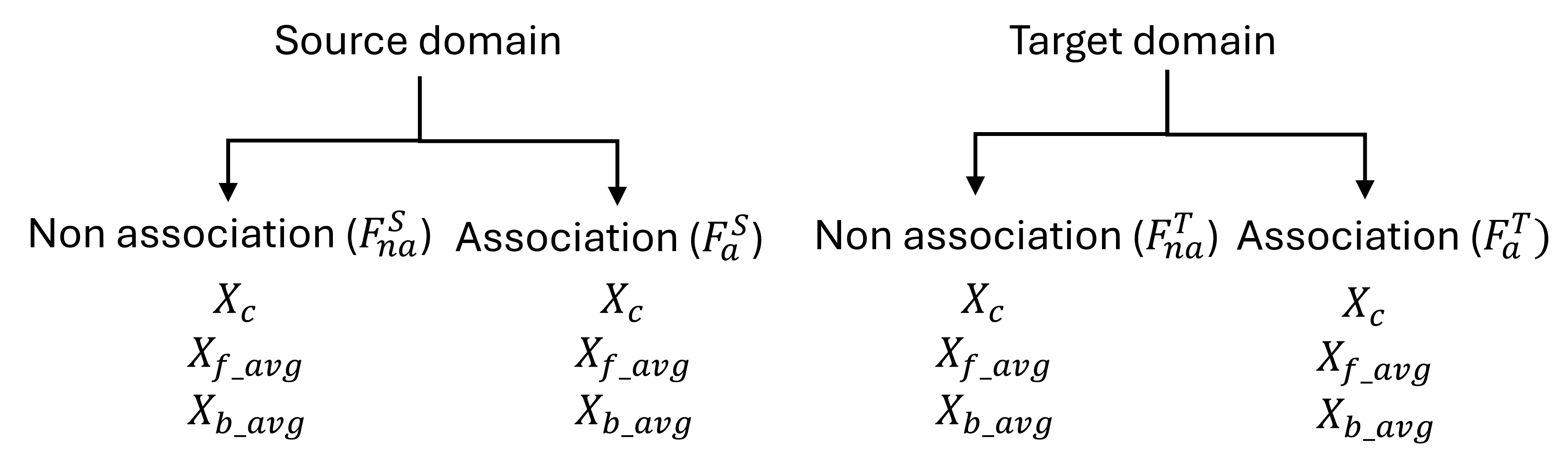}
\end{center}
\caption{\textbf{Feature extraction graph.} We extracted features from all true positive cases with different FG-BG pairs.}
\label{fig:extract features graph}
\end{figure}

The results from our experiments confirmed the existence and causal effect of FG–BG associations. However, these findings did not directly quantify its influence on DAOD. To address this, we introduced a new metric, defined as domain association gradient ($Gradient$), which measures performance perturbation in response to FG–BG associations strength. For the association strength, we add the MMD between FG and BG and subtract the MMD in FG and BG. Figure \ref{fig:extract features graph} illustrates the feature extraction breakdown. For each domain ($D$), we can separate the features into association ($F^D_a$) and non-association features ($F^D_{na}$). These features can then be further broken down into CAM activated features ($X_c$), 2D pooled averaged FG features within $X_c$ ($X_{f\_avg}$) and 2D pooled averaged BG features within $X_c$ ($X_{b\_avg}$) using instance masks to make FG and BG separable. The detail process to extract each feature from FG and BG of different domains (e.g. $X_c$, $X_{f\_avg}$, $X_{b\_avg}$) is described in Section \ref{sec:ass no-ass features}.

The intuition is that if a strong FG–BG association is learned, the MMD between features that preserve this association should be small, due to shared contextual dependencies. Besides, drop rate should be large with a strong FG-BG associations. Consequently, the $Gradient$ value that computed using these features will be larger than those derived from features lacking FG–BG associations. We calculated $Gradient$ values using the drop rate for each FG–BG pair and analyzed their patterns via statistical testing using equations~\ref{eq:gradient_equations}. Only FG–BG pairs with statistically significant drop rate differences identified in Experiment~\ref{exp2:FG-BG drop rate feature space} were included in this analysis. Based on our hypothesis, we expected $\text{Gradient}^S$ to exceed $\text{Gradient}^T$, as models trained on the source domain tend to capture stronger FG–BG associations. Each superscript ($S$ and $T$) denotes the source and target domain, respectively. 

\begin{equation}
\begin{aligned}
\text{Gradient}^{\text{S}} &= \frac{\text{Source domain drop rate}}{\text{f2b}^{\text{S}} + \text{b2f}^{\text{S}} - \text{f2f}^{\text{S}} - \text{b2b}^{\text{S}}}, \quad
\text{Gradient}^{\text{T}} &= \frac{\text{Target domain drop rate}}{\text{f2b}^{\text{T}} + \text{b2f}^{\text{T}} - \text{f2f}^{\text{T}} - \text{b2b}^{\text{T}}}
\end{aligned}
\label{eq:gradient_equations}
\end{equation}

Equation~\ref{eq:mmd_metrics} defines how various feature combinations, drawn from different contexts, are used in computing this metric. Specifically, cross-context feature comparisons, such as \texttt{f2b} and \texttt{b2f}, are expected to yield higher MMD values than within-context comparisons like \texttt{f2f} and \texttt{b2b}, since the latter reflect similar contextual structures.

\begin{equation}
\begin{aligned}
\text{f2f}^{\text{S}} &= \text{MMD}(X_{f\_avg} \in F^S_{na},\, X_{f\_avg} \in F^S_{a}) &\quad
\text{f2f}^{\text{T}} &= \text{MMD}(X_{f\_avg} \in F^T_{na},\, X_{f\_avg} \in F^T_{a}) \\
\text{f2b}^{\text{S}} &= \text{MMD}(X_{f\_avg} \in F^S_{na},\, X_{b\_avg} \in F^S_{a}) &\quad
\text{f2b}^{\text{T}} &= \text{MMD}(X_{f\_avg} \in F^T_{na},\, X_{b\_avg} \in F^T_{a}) \\
\text{b2f}^{\text{S}} &= \text{MMD}(X_{b\_avg} \in F^S_{na},\, X_{f\_avg} \in F^S_{a}) &\quad
\text{b2f}^{\text{T}} &= \text{MMD}(X_{b\_avg} \in F^T_{na},\, X_{f\_avg} \in F^T_{a}) \\
\text{b2b}^{\text{S}} &= \text{MMD}(X_{b\_avg} \in F^S_{na},\, X_{b\_avg} \in F^S_{a}) &\quad
\text{b2b}^{\text{T}} &= \text{MMD}(X_{b\_avg} \in F^T_{na},\, X_{b\_avg} \in F^T_{a})
\end{aligned}
\label{eq:mmd_metrics}
\end{equation}
Furthermore, we conducted a focused analysis on \texttt{f2b} and \texttt{b2f} MMD values for both associated and non-associated features across domains using equation \ref{eq:mmd_cross_domain} and \ref{eq:sum_mmd_components}. We computed the sum of these values and compared them using pairwise T-test \citep{o2017sensory} after Shapiro-Wilk test \citep{shapiro1965analysis}. 

\begin{equation}
\begin{aligned}
\text{$f2b_{asso}$} &= \text{MMD}(X_{f\_avg} \in F^S_{a},\, X_{b\_avg} \in F^T_{a}) &\quad
\text{$f2b_{no-asso}$} &= \text{MMD}(X_{f\_avg} \in F^S_{na},\, X_{b\_avg} \in F^T_{na}) \\
\text{$b2f_{asso}$} &= \text{MMD}(X_{b\_avg} \in F^S_{a},\, X_{f\_avg} \in F^T_{a}) &\quad
\text{$b2f_{no-asso}$} &= \text{MMD}(X_{b\_avg} \in F^S_{na},\, X_{f\_avg} \in F^T_{na}) \\
\end{aligned}
\label{eq:mmd_cross_domain}
\end{equation}
\begin{equation}
\begin{aligned}
\text{sum}_{\text{a-a}} &= \text{f2b}_{\text{asso}} + \text{b2f}_{\text{asso}} & \quad
\text{sum}_{\text{na-na}} &= \text{f2b}_{\text{no-asso}} + \text{b2f}_{\text{no-asso}}
\end{aligned}
\label{eq:sum_mmd_components}
\end{equation}
A higher \texttt{$sum_{na-na}$} than \texttt{$sum_{a-a}$} indicates that FG–BG association persists across domains and may influence DAOD performance when the model unintentionally leverages such cross-domain dependencies during inference. In other words, FG-BG association  was not impact on across DAOD then \texttt{$sum_{na-na}$} and \texttt{$sum_{a-a}$} should be no statistical significance different.
We randomly sampled non-associated features to correspond to the number of associated features. We divided all datasets into two groups. The first group includes Cityscapes related datasets with ``KST'' and the second group contains only Virtual KITTI related dataset due to unmatched label configurations. 

\subsubsection{Associated Features and Non-associated Features Extraction}
\label{sec:ass no-ass features}
With drop rate experiment (Algorithm \ref{algo:CAM FG-BG association}) as pre-processing, we defined associated features when the extracted features of each instance did not detect without background. In other words, it indicates the features encode association between FG-BG. Conversely, non-associated features indicate that FG-BG association is not included in the extracted feature. We extracted features and clustered per FG-BG pair. We used 4 different layers for each model to extract features in different scales. For ``Res'' and ALDI++, we used \texttt{$res2.2, res3.3, res4.5, res5.2$} and for ``Eff'', we used \texttt{$backbone.block.0, 1, 3$}, and   \texttt{$5$}. From each domain, we saved associated features and non-associated features. 
Algorithm~\ref{algo:feature_extraction} describes steps to process foreground-related features and background-related features. Table~\ref{tab:association_no-association table} defines $F^D_a$ and $F^D_{na}$ meaning FG-BG associated features and FG-BG non-associated features from each domain $D$.

\begin{center}
\begin{minipage}{0.9\linewidth}
\begin{algorithm}[H]
\caption{Feature Extraction from CAM and Ground Truth Instance Mask}
\label{algo:feature_extraction}
\KwIn{
    \begin{itemize}
        \item $C$: CAM mask from Algorithm \ref{algo:CAM FG-BG association}.
        \item $G$: Ground truth instance binary mask
        \item $A$: Activation maps from different layers (e.g. \texttt{$res2.2, res3.3, res4.5$}, and \texttt{$res5.2$} for ``Res'')
    \end{itemize}
}
\KwOut{$X_{c}$, $X_{f\_avg}$, $X_{b\_avg}$}

Compute features $X$ from CAM mask $C$\;
    \Indp $X$ = $A \cdot \mathbbm{1}(C = 1)$\;
    \Indm
Compute Normalized features $X_{c}$ from $X$\;
\Indp $X_{c}$ =\text{Normalize}($X$)\;
\Indm

Separate normalized features using ground truth mask $G$: \\
\Indp $X_{f\_avg}$ =  \text{Adaptive pool 2d}($X_{c} \cdot \mathbbm{1}(G = 1)$)\;
$X_{b\_avg}$ = \text{Adaptive pool 2d}($X_{c} \cdot \mathbbm{1}(G = 0)$)\;
\Indm
\end{algorithm}
\end{minipage}
\end{center}

\begin{table}[h!]
\centering
\caption{\textbf{Definition of FG-BG associated features and non-associated features at each domain $D$.} }
\label{tab:association_no-association table}
\small
\renewcommand{\arraystretch}{1.2}
\begin{tabular}{|c|>{\centering\arraybackslash}m{1.2cm}|m{5cm}|m{5cm}|}
\hline
\multicolumn{2}{|c|}{\multirow{2}{*}{}} & \multicolumn{2}{c|}{\textbf{Detection}} \\
\cline{3-4}
\multicolumn{2}{|c|}{} & \multicolumn{1}{c|}{\textbf{0}} & \multicolumn{1}{c|}{\textbf{1}} \\
\hline
\multirow{2}{*}{\textbf{BG removal}} & \textbf{1} & 
When ``road'' is removed, detection fails $\rightarrow$ car feature with association ($F^D_a$) & 
Without ``road'', detection succeeds $\rightarrow$ car feature without association ($F^D_{na})$\\
\cline{2-4}
& \textbf{0} & 
False Negative. Unknown association impact on prediction. & 
True Positive. No association impact on prediction. \\
\hline
\end{tabular}
\end{table}

\section{Experiments}
\subsection*{Q1. Are FG-BG associations being inadvertently learned during the training process?}
\subsubsection*{Model Evaluation}
We evaluated the trained models using mAP@50 metric. Table \ref{tab:model_evalaution} summarizes the evaluation results. Among models trained on the Cityscapes dataset, ALDI++ outperformed others on Cityscapes-related datasets and ``KST'', likely due to longer training epochs on target domain and the use of multiple domain datasets via DAOD algorithms. Although ``Yo'' achieved strong performance on ``CST'', its domain adaptation capability was weaker than that of the baseline ``Res''. ``Eff'' consistently showed the lowest performance on both Cityscapes and ``KST'' but outperformed ``Yo'' on ``KST''. For models trained on ``KST'', ALDI++ outperformed other models, while ``Eff'' and ``Yo'' showed significantly poor results. ``KST'' dataset, with only 200 images, introduced a domain shift that limited generalization to larger datasets. Consequently, ``Res'' and ``Yo'' also exhibited weakened performance due to insufficient data. With VKC-trained models, ``Res'' and ``Yo'' demonstrated reasonable domain generalization compared to ``Eff''. However, on the ``VKF'' validation set, ``Yo''’s performance dropped significantly relative to ``Res'', despite its overall strong results on other datasets. ``Eff'' also experienced a notable performance decline on ``VKF''. ALDI++ demonstrated the effectiveness of domain generalization methods.

\begin{table*}[hbt!]
\centering
\caption{\textbf{Model evaluation across different training and validation sets.} ``-'' is not measurable.}
\resizebox{\textwidth}{!}{%
\begin{tabular}{l|cccc|cccc|cccc}
\hline
\textbf{Dataset} 
& \multicolumn{4}{c|}{\textbf{Cityscapes Train}} 
& \multicolumn{4}{c|}{\textbf{KST  Train}} 
& \multicolumn{4}{c}{\textbf{VKC Train}} \\
 & \textbf{Res} & \textbf{Eff} & \textbf{Yo} & \textbf{ALDI++}
 & \textbf{Res} & \textbf{Eff} & \textbf{Yo} & \textbf{ALDI++}
 & \textbf{Res} & \textbf{Eff} & \textbf{Yo} & \textbf{ALDI++} \\
\hline
CST  & 79.14 & 41.12 & \textbf{88.26} & 87.97 & -     & -     & -   & - & -     & -               & - & -\\
CSV  & 67.59 & 42.90 & 59.56& \textbf{70.08} & 43.23 & 2.74  & 21.69& \textbf{51.09} & - & -  & - &-\\
CFV & 57.13 & 20.58 & 44.49& \textbf{67.45} & 35.48 & 0.61  & 12.53& \textbf{43.62} & - & - & - &-\\
CRV & 58.65 & 23.18 & 48.77& \textbf{69.78} & 37.42 & 0.86  & 17.57& \textbf{47.82} & - & - & - &-\\
KST & 46.25 & 28.92 & 23.91& \textbf{47.96} & 86.17 & 10.42 & 21.53& \textbf{92.44} & - & -   & - &-\\
VKC & - & - & - & - & - &  - & - & - & 81.67 & 50.09   & \textbf{85.99}       & 81.96 \\
VKF & - & - & - & - & - &  - & - & - & 61.14 & 5.80           &34.27& \textbf{72.60} \\
VKM & - & - & - & - & - &  - & - & - & 79.72 & 29.52   & 79.55 & \textbf{80.27} \\
VKO & - & - & - & - & - &  - & - & - & 75.14 & 30.41   & \textbf{81.93}       & 78.58 \\
VKR & - & - & - & - & - &  - & - & - & 71.66 & 25.53   & 75.93       & \textbf{78.36} \\
VKS & - & - & - & - & - &  - & - & - & 76.02 & 26.18   & \textbf{78.65}       & 77.59 \\
\hline
\end{tabular}
}
\label{tab:model_evalaution}
\end{table*}

\subsubsection*{Q1-Exp1. Class-wise Background Removal Experiments in Image Space}
This experiment evaluated the role of FG-BG associations by replacing background regions with non-salient object images while preserving foreground objects. We measured mAP@50 over six repetitions using randomly generated images. Table~\ref{tab:bg_image_space_evaluation} presents the mean and standard deviation across six evaluation runs. Models trained on Cityscapes showed substantial performance drops compared to their original evaluation. Notably, ALDI++ exhibited greater degradation than ``Res'' on ``KST'', suggesting that ALDI++ strongly relies on FG-BG associations learned from both source and target domains. Similarly, models trained on ``KST'' and ``VKC'' also experienced considerable performance declines, indicating that FG-BG associations were learned during training and utilized during inference. 

\begin{table*}[hbt!]
\centering
\caption{\textbf{Mean $\pm$ standard deviation of mAP@50 across synthetic datasets and models trained on Cityscapes, ``KST'', and ``VKC''.} The bolded values are the highest mAP@50 for each train dataset-model pair.}
\label{tab:bg_image_space_evaluation}
\begin{center}

\begin{minipage}{0.9\textwidth}
\centering
\textbf{Cityscapes Trained} \\
\vspace{0.2em}
\begin{tabular}{c|cccc}
\hline
\textbf{Dataset + BG} & Res & Eff & Yo & ALDI++ \\
\hline
CSV & 44.5 $\pm$ 6.1  & 10.02 $\pm$ 6.4 & 32.78 $\pm$ 0.6 & \textbf{47.96} $\pm$ 6.3 \\
CFV & 34.6 $\pm$ 10.6 & 8.99 $\pm$ 4.5 & 15.70 $\pm$ 0.3 & \textbf{43.17} $\pm$ 9.8 \\
CRV & 29.1 $\pm$ 4.4  & 14.24 $\pm$ 8.4 & 17.37 $\pm$ 0.3 & \textbf{38.18} $\pm$ 5.1 \\
KST & \textbf{41.4} $\pm$ 4.4 & 25.33 $\pm$ 6.4 & 19.92 $\pm$ 1.4 & 33.98 $\pm$ 6.7 \\
\hline
\end{tabular}
\end{minipage}

\par\vspace{1em} 

\begin{minipage}{0.9\textwidth}
\centering
\textbf{KST Trained} \\
\vspace{0.2em}
\begin{tabular}{c|cccc}
\hline
\textbf{Dataset + BG} & Res & Eff & Yo & ALDI++ \\
\hline
CSV & 19.2 $\pm$ 5.2 & 0.16 $\pm$ 0.1 & 21.34 $\pm$ 0.3 & \textbf{23.01} $\pm$ 0.7 \\
CFV & 17.0 $\pm$ 5.8 & 0.22 $\pm$ 0.1 & 10.35 $\pm$ 0.1 & \textbf{24.55} $\pm$ 0.1 \\
CRV & 13.6 $\pm$ 3.4 & 2.25 $\pm$ 2.1 & 14.29 $\pm$ 0.1 & \textbf{35.29} $\pm$ 0.7 \\
\hline
\end{tabular}
\end{minipage}
\par\vspace{1em}
\begin{minipage}{0.9\textwidth}
\centering
\textbf{VKC Trained} \\
\vspace{0.2em}
\begin{tabular}{c|cccc}
\hline
\textbf{Dataset + BG} & Res & Eff & Yo & ALDI++ \\
\hline
VKF & 38.7 $\pm$ 10.7 & 1.17 $\pm$ 0.5 & 14.61 $\pm$ 0.5  & \textbf{51.04} $\pm$ 0.5\\
VKM & 61.8 $\pm$ 5.1  & 11.44 $\pm$ 6.4 & 35.26 $\pm$ 0.6 & \textbf{64.17} $\pm$ 0.3\\
VKO & 60.0 $\pm$ 4.7  & 10.70 $\pm$ 5.5 & 33.66 $\pm$ 0.4 & \textbf{64.44} $\pm$ 0.5\\
VKR & 58.4 $\pm$ 3.5  & 7.99 $\pm$ 3.5 & 27.56 $\pm$ 0.4  & \textbf{59.81} $\pm$ 0.4\\
VKS & 59.8 $\pm$ 6.5  & 12.92 $\pm$ 4.6 & 35.93 $\pm$ 0.8 & \textbf{64.26} $\pm$ 0.2\\
\hline
\end{tabular}
\end{minipage}
\end{center}
\end{table*}

\subsubsection*{Q1-Exp2. Feature-wise Background Removal Experiments in Feature Space}
\begin{figure}[h]
\begin{center}
\centering
\includegraphics[width=0.9\textwidth]{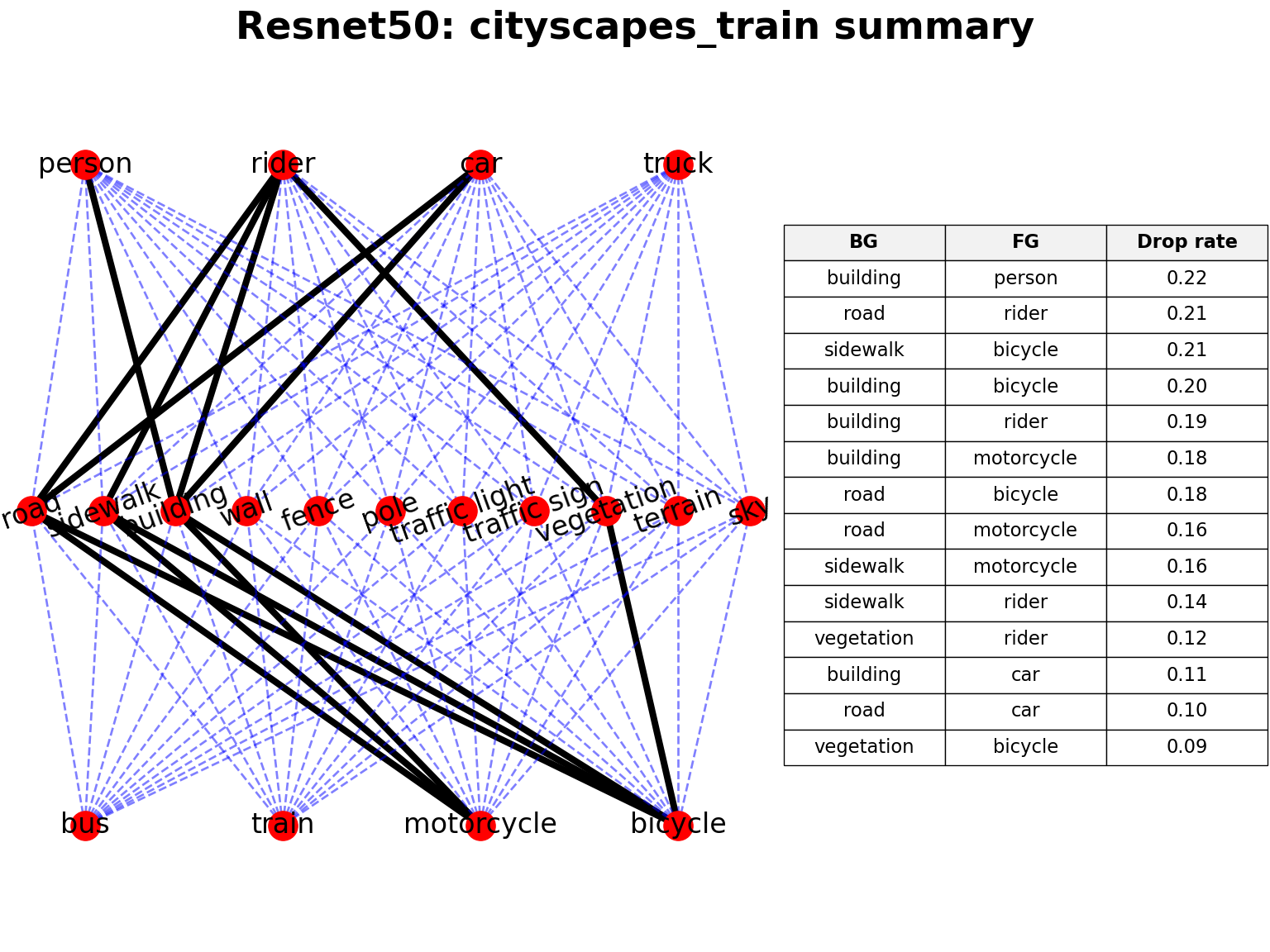}
\end{center}
\caption{\textbf{Feature-wise drop rates of ``Res'' on ``CST''.} Bold edges indicates statistically significant difference on the FG-BG pair. The table on the right in the figure illustrates the drop rate in decreasing order.}
\label{fig:feature-wise drop rate}
\end{figure}

In addition to background perturbation in image space, we performed background removal in feature space in Table~\ref{tab:drop_rate_feature_space}. Specifically, we zeroed out activated background regions in the shallow layers of each model architecture, preventing background information from propagating to deeper layers. This effectively disables the FG-BG association during inference. The Cityscapes-related datasets and ``KST'' contain 88 distinct FG-BG combinations, while the Virtual KITTI datasets include 30 combinations. We report only the combinations that resulted in a statistically significant performance drop of at least 8\%. Note that additional combinations exhibited smaller drops and are not included in the table. Figure \ref{fig:feature-wise drop rate} illustrates the example of significant performance drop on ``CST'' with ``Res'' model. It indicates that the models learned notable FG-BG associations during training, which enable cause performance degradation.

\begin{table}[h]
\centering
\caption{\textbf{The number of FG-BG pairs statistically significant different across models}. Only FG-BG pairs more than 8\% drop rate are denoted. The bold values indicate stronger FG–BG associations for each model across the respective datasets.}
\begin{tabular}{|c|c|c|c|c|}
\hline
\textbf{} & \textbf{Res} & \textbf{ALDI++} & \textbf{Eff} & \textbf{Yo} \\
\hline
CST  & 14/88 & \textbf{18/88} & 15/88 & 12/88 \\
CSV  & 7/88  & \textbf{17/88} & 7/88  & 3/88 \\
CFV  & 13/88 & \textbf{20/88} & 11/88 & 7/88 \\
CRV  & 15/88 & \textbf{21/88} & 12/88 & 6/88 \\
KST  & 2/88  & 2/88  & \textbf{4/88}  & 2/88\\
VKC  & 8/30  & 7/30   & \textbf{9/30} & 4/30\\
VKF  & \textbf{7/30}  &  \textbf{7/30}    & 4/30 & 2/30\\
VKM  & \textbf{9/30}  &  7/30    & 7/30  & 3/30\\
VKO  & \textbf{9/30}  &  7/30    & 8/30  & 3/30 \\
VKR  & 7/30  &  \textbf{9/30}    & \textbf{9/30} & 6/30\\
VKS  & 8/30  &  8/30      & \textbf{9/30} & 4/30\\
\hline
\end{tabular}
\label{tab:drop_rate_feature_space}
\end{table}

\subsection*{Q2. Is FG-BG Associations Causal Relationship with Object Detection?}
In the previous section, we validated the existence of FG-BG associations. To further investigate this phenomenon, we computed CAM masks for each object and analyzed background-region-based associations. To explore the causal relationship, we applied do-calculus using CAM-derived masks and ground truth instance masks. We also note that ``Yo'' contains non-differentiable non-maximum suppression (NMS) which stops us from using gradient based CAM to capture specific objects' contextual masks. Thus, we are not able to answer Q2 and Q3 experiments using this particular model.

\subsubsection*{Q2-Exp1. FG-BG Association with Respect to Activated Background Region}
With do-calculus, we computed mean of drop rate of all classes per bin. While foreground region were maintained regardless of bin (see Fig. \ref{fig:resnet50_bins} and Table \ref{tab:bin_hit_ratio}), the drop rate significantly increased with bin 1 which has small amount of activated background (see Table \ref{tab:drop_rate_bin}). While enlarging the background regions, the drop rate converged to 0.0 which means the objects were detected correctly. 
Through the experiment, we confirmed a causal relationship between FG-BG associations and outcomes, as the accuracy of foreground objects changed notably across bins, particularly across from bin 1 to bin 5.
\begin{table}[h!]
\centering
\caption{\textbf{Definition of FG-BG associated features and non-associated features for each domain D.} The FG mean of 1.0 indicates that all foreground pixels are captured by the CAM. The BG mean represents the ratio of captured background pixels to the total number of foreground pixels, reflecting the extent of background activation relative to the foreground.}
\label{tab:bin_hit_ratio}
\begin{tabular}{|l|c|c|c|c|}
\hline
\multirow{2}{*}{Layer} & \multicolumn{4}{c|}{Hit ratio} \\
\cline{2-5}
\multirow{2}{*}{} & \multicolumn{2}{c|}{Associated} & \multicolumn{2}{c|}{Non-associated}  \\
\cline{2-5}
                       & FG mean     & BG mean     & FG mean     & BG mean     \\
\hline
\texttt{$res2.2$} & 1.0 & 14.81 & 1.0  & 20.04  \\
\texttt{$res3.3$} & 1.0 & 14.33 & 1.0  & 18.79  \\
\texttt{$res4.5$} & 1.0 & 12.66 & 1.0  & 27.2  \\
\texttt{$res5.2$} & 1.0 & 21.0  & 1.0  & 12.33\\
\hline
\end{tabular}
\end{table}

\begin{table}[htbp]
\centering
\caption{\textbf{Drop rate per bin.} 5 numbers in each cell are bin 1, 3, 5, 7, and 9 (``B''). The lower means fewer or no drop rates were measured. The bolded values highlight the significant performance drop.}
\begin{adjustbox}{max width=\textwidth}
\begin{tabular}{|l|ccccc|ccccc|ccccc|}
\hline
\textbf{} & \multicolumn{5}{c|}{\textbf{Res}} & \multicolumn{5}{c|}{\textbf{ALDI++}} & \multicolumn{5}{c|}{\textbf{Eff}} \\
\hline
          & B1 & B3 & B5 & B7 & B9 & B1 & B3 & B5 & B7 & B9 & B1 & B3 & B5 & B7 & B9\\
\hline
CST  & \textbf{0.65} & 0.14 & 0.02 & 0.01 & 0.00 & \textbf{0.66} & 0.18 & 0.02 & 0.00 & 0.00 & \textbf{0.74} & 0.31 & 0.25 & 0.26 & 0.18 \\
CSV  & \textbf{0.62} & 0.10 & 0.02 & 0.00 & 0.00 & \textbf{0.64} & 0.11 & 0.01 & 0.00 & 0.00 & \textbf{0.71} & 0.18 & 0.14 & 0.13 & 0.10 \\
CFV  & \textbf{0.67} & 0.12 & 0.03 & 0.00 & 0.00 & \textbf{0.68} & 0.10 & 0.02 & 0.01 & 0.00 & \textbf{0.79} & 0.19 & 0.09 & 0.11 & 0.11 \\
CRV  & \textbf{0.67} & 0.27 & 0.07 & 0.02 & 0.01 & \textbf{0.64} & 0.20 & 0.05 & 0.01 & 0.01 & \textbf{0.65} & 0.12 & 0.06 & 0.07 & 0.09 \\
KST  & \textbf{0.84} & 0.53 & 0.22 & 0.01 & 0.00 & \textbf{0.70} & 0.31 & 0.00 & 0.02 & 0.02 & \textbf{0.82} & 0.71 & 0.36 & 0.69 & 0.68 \\
VKC  & \textbf{0.57} & 0.08 & 0.02 & 0.00 & 0.00 & \textbf{0.33} & 0.03 & 0.01 & 0.00 & 0.00   & \textbf{0.71} & 0.19 & 0.07 & 0.04 & 0.03 \\
VKF  & \textbf{0.75} & 0.37 & 0.12 & 0.04 & 0.01 & \textbf{0.42} & 0.05 & 0.02 & 0.01 & 0.00    & \textbf{0.92} & 0.15 & 0.04 & 0.05 & 0.02 \\
VKM  & \textbf{0.63} & 0.10 & 0.03 & 0.00 & 0.00 & \textbf{0.42} & 0.05 & 0.02 & 0.01 & 0.00    & \textbf{0.77} & 0.25 & 0.11 & 0.06 & 0.05 \\
VKO  & \textbf{0.56} & 0.11 & 0.03 & 0.00 & 0.00 & \textbf{0.33} & 0.05 & 0.02 & 0.00 & 0.00   & \textbf{0.76} & 0.23 & 0.05 & 0.03 & 0.05 \\
VKR  & \textbf{0.62} & 0.13 & 0.04 & 0.01 & 0.00 & \textbf{0.33} & 0.05 & 0.02 & 0.00 & 0.00   & \textbf{0.80} & 0.26 & 0.11 & 0.08 & 0.07 \\
VKS  & \textbf{0.64} & 0.16 & 0.07 & 0.02 & 0.01 & \textbf{0.34} & 0.05 & 0.02 & 0.00 & 0.00   & \textbf{0.77} & 0.21 & 0.06 & 0.03 & 0.03 \\
\hline
\end{tabular}
\label{tab:drop_rate_bin}
\end{adjustbox}
\end{table}

\subsection*{Q3. Is FG-BG Associations Impact on DAOD and How to Quantify the Effect? }
We computed $Gradient$ values and analyzed the MMDs between associated and non-associated features across different domains. This allowed us to quantify the impact of FG-BG associations on domain shifts.

\subsubsection*{Q3-Exp1. Domain Association Gradient}
\begin{figure}[h]
\begin{center}
\centering
\includegraphics[width=0.9\textwidth]{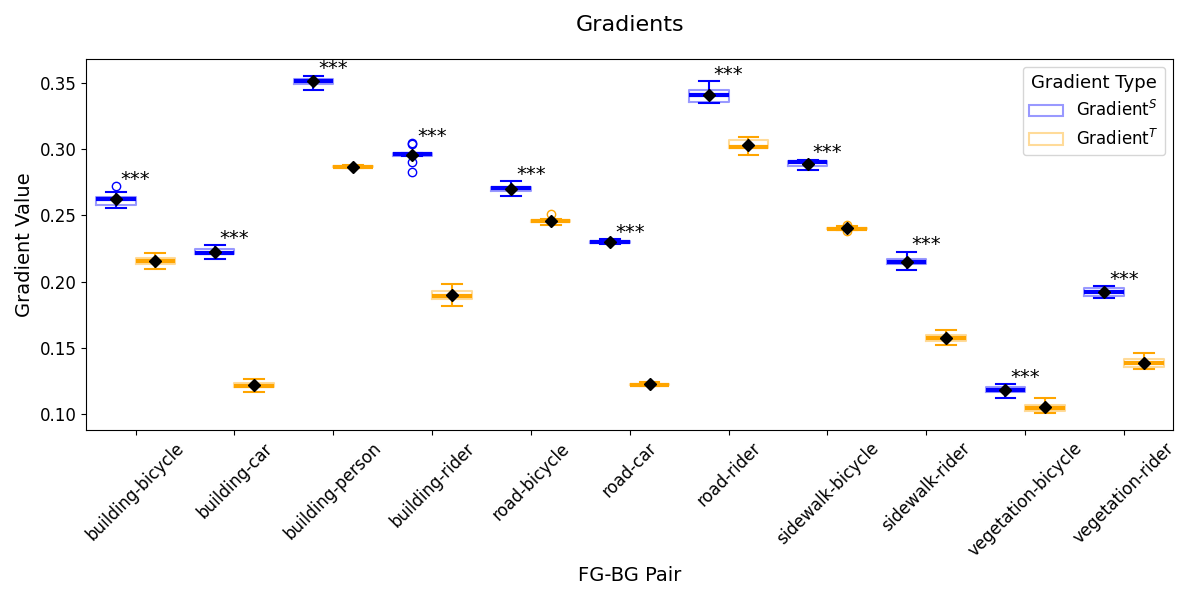}
\end{center}
\caption{$\boldsymbol{Gradient^S}$\textbf{ and }$\boldsymbol{Gradient^T}$\textbf{ comparison}. The number of ``*'' indicate the level of statistical significance of a p-value.}
\label{fig:gradients}
\end{figure}

To validate our hypothesis that $Gradient^S$ should be lower than $Gradient^T$ due to learned FG-BG associations, we categorized the comparisons into three cases: \textbf{(1) $\boldsymbol{Gradient^S}$ significantly lower than $\boldsymbol{Gradient^T}$, (2) $\boldsymbol{Gradient^S}$ significantly higher than $\boldsymbol{Gradient^T}$}, and \textbf{(3) no statistically significant difference.} Table~\ref{tab:Gradient_comparison} summarizes the results. Overall, the findings support our hypothesis. However, ALDI++ showed opposite results on the CST-CSV pair, possibly due to DAOD training strategy using target domain information (CFV). In the Virtual KITTI-related datasets, particularly for ``Eff'', some results contradicted expectations. This may be attributed to strong spatial and temporal correlations inherent in the dataset, which is derived from object-tracking video sequences, or to biases introduced by a small number of detections and drops. These results elaborates quantification of class-wise causal effects in DAOD. Figure \ref{fig:gradients}  presents the results of the $Gradient$ comparisons as box plot to present the visual comparison.

\begin{table}[htbp]
\centering
\caption{\textbf{$Gradient$ comparison across different domains.} Each number's denominator is the number of FG-BG association in common across two domains. Cases 1, 2 and 3 are given as C1, C2 and C3. The bolded values represent the dominant case for each dataset-model pair. }
\label{tab:gradients}
\begin{tabular}{|l|ccc|ccc|ccc|}
\hline
\textbf{} & \multicolumn{3}{c|}{\textbf{Res}} & \multicolumn{3}{c|}  {\textbf{ALDI++}} & \multicolumn{3}{c|}{\textbf{Eff}} \\
\cline{2-10}
          & C1 & C2 & C3 &  C1 & C2 & C3 &  C1 & C2 & C3 \\
\hline
CST - CSV   & \textbf{4/5  }& 0/5  & 1/5  & 4/10 & \textbf{6/10} & 0/10 & \textbf{3/7}  & \textbf{3/7}  & 1/7  \\
CST - CFV   & \textbf{11/11} & 0/11 & 0/11 & \textbf{15/15} & 0/15 & 0/15 & \textbf{7/11} & 2/11 & 2/11 \\
CST - CRV   &1\textbf{0/11 }& 1/11 & 0/11 & \textbf{8/10} & 2/10 & 0/10 & \textbf{7/10} & 3/10 & 0/10 \\
CST - KST   & \textbf{1/1  }& 0/1  & 0/1  &  \textbf{1/1  }& 0/1  & 0/1   & \textbf{2/3}  & 1/3  & 0/3  \\
VKC - VKF   & \textbf{5/5  }& 0/5  & 0/5  &  \textbf{4/4  }& 0/4  & 0/4   & \textbf{1/2}  & \textbf{1/2}  & 0/2  \\
VKC - VKM   & \textbf{5/7  }& 2/7  & 0/7  &  \textbf{4/4  }& 0/4  & 0/4   & 1/5  & \textbf{4/5}  & 0/5  \\
VKC - VKO   & \textbf{4/7  }& 3/7  & 0/7  &  \textbf{4/4  }& 0/4  & 0/4   & 1/5  & \textbf{4/5}  & 0/5  \\
VKC - VKR   & \textbf{6/6  }& 0/6  & 0/6  &  \textbf{4/5  }& 1/5  & 0/5   & 1/5  & \textbf{4/5}  & 0/5  \\
VKC - VKS   & 2/6  & \textbf{4/6}  & 0/6  &  \textbf{4/5 }& 1/5  & 0/5   & 1/5  & \textbf{4/5}  & 0/5  \\
\hline
\end{tabular}
\label{tab:Gradient_comparison}
\end{table}

\subsubsection*{Q3-Exp2. Associated and Non-associated Features}
\begin{figure}[h]
\begin{center}
\centering
\includegraphics[width=0.9\textwidth]{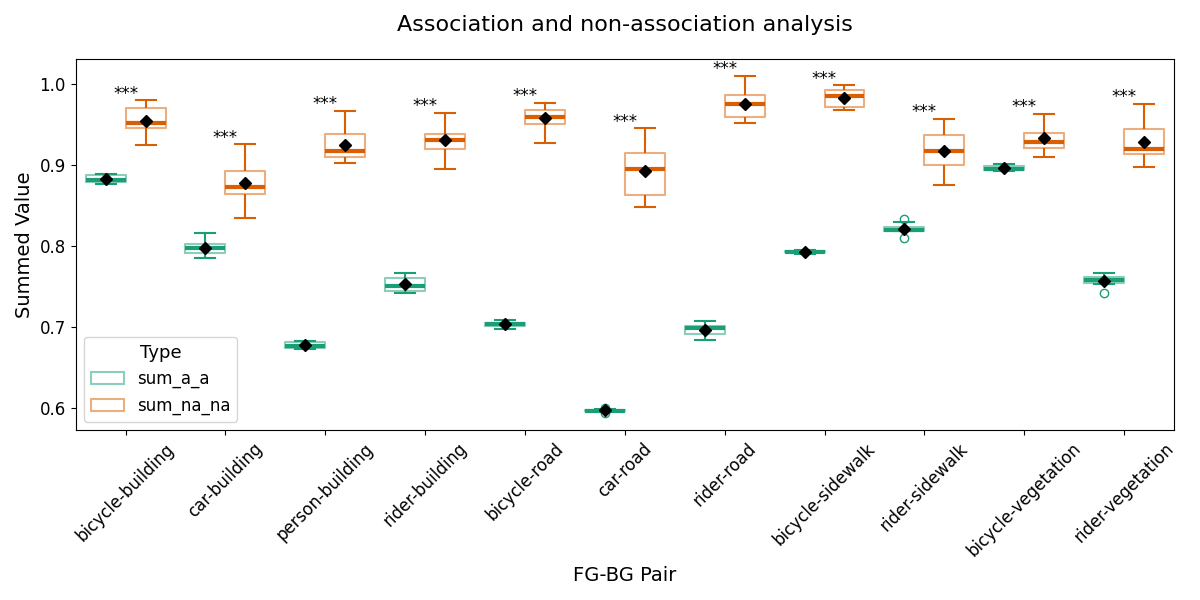}
\end{center}
\caption{\textbf{MMD of associated features and non-associated comparison}. The number of ``*'' indicate the level of statistical significance of a p-value.}
\label{fig:aa_nana}
\end{figure}

To understand the class- and feature-wise impact across different domains, we compared the gap between the MMD of associated features and that of non-associated features. Similar to previous $Gradient$ analysis,  we categorized the comparisons into three cases: \textbf{(1) The summation of MMD of \texttt{f2b} and \texttt{b2f} in associated features significantly lower than that of non-associated features (2) The summation of MMD of \texttt{f2b} and \texttt{b2f} in associated features significantly higher than that of non-associated features}, and \textbf{(3) no statistically significant difference.} Overall, associated features sharing the same FG-BG association across domains exhibited lower MMD than non-associated features, indicating stronger FG-BG association consistency in cross-domain. However, ``Eff'' showed reversed outcomes on the CST-CRV and CST-KST pairs. This may be due to a limited number of detections, resulting in insufficient feature representations or overall poor model performance. Table~\ref{tab:aa_nana_comparison} presents the results, while Figure~\ref{fig:aa_nana} depicts the results of associated and non-associated feature comparison, with statistical analysis annotated in the box plots.

\begin{table}[h]
\centering
\caption{\textbf{Associated and non-associated features comparison across different domains.} Each number's denominator is the number of FG-BG association in common across two domains. Cases 1, 2 and 3 are given as C1, C2 and C3. ``-'' is not measurable statistically. The bolded values represent the dominant case for each dataset-model pair.}
\begin{tabular}{|l|ccc|ccc|ccc|}
\hline
\textbf{} & \multicolumn{3}{c|}{\textbf{Res}} & \multicolumn{3}{c|}  {\textbf{ALDI++}} & \multicolumn{3}{c|}{\textbf{Eff}} \\
\cline{2-10}
          & C1 & C2 & C3 &  C1 & C2 & C3 &  C1 & C2 & C3 \\
\hline
CST - CSV  & \textbf{5/5  } & 0/5     & 0/5     & \textbf{9/10}    & 0/10& 1/10   & \textbf{3/7}& 2/7& 2/7   \\
CST - CFV  & \textbf{11/11} & 0/11    & 0/11    & \textbf{9/15}    & 6/15& 0/15   & \textbf{6/11}& 5/11& 0/11 \\
CST - CRV  & \textbf{8/11 } & 3/11    & 0/11    & \textbf{7/10}    & 3/10& 0/10   & 2/10& \textbf{6/10}& 2/10 \\
CST - KST  & \textbf{1/1} & 0/1     & 0/1     & \textbf{1/1} & 0/1     & 0/1  & 0/3& \textbf{2/3}& 1/3   \\
VKC - VKF  & \textbf{2/5} & 1/5     & \textbf{2/5}     & \textbf{4/4} & 0/4 & 0/4 & \textbf{2/2}& 0/2& 0/2   \\
VKC - VKM  & \textbf{4/7} & 2/7     & 1/7     &  \textbf{4/4} & 0/4 & 0/4 & \textbf{3/5} & 1/5& 1/5   \\
VKC - VKO  & \textbf{5/7} & 2/7     & 0/7     & \textbf{4/4} & 0/4 & 0/4 & \textbf{4/5} & 1/5& 1/5   \\
VKC - VKR  & \textbf{4/6} & 1/6     & 1/6     & \textbf{5/5} & 0/5 & 0/5  & \textbf{4/5} & 1/5& 0/5   \\
VKC - VKS  & \textbf{5/6} & 1/6     & 0/6     & \textbf{5/5} & 0/5 & 0/5  & \textbf{3/5} & 1/5& 1/5   \\
\hline
\end{tabular}
\label{tab:aa_nana_comparison}
\end{table}

\section{Discussion and Conclusion}
\label{sec:discussion}
In this work, we present a comprehensive empirical and theoretical investigation into the role of context bias in domain-adaptive object detection (DAOD). While context bias has previously been studied in classification and segmentation tasks, our work is the first to formally identify, quantify, and causally analyze this phenomenon within the context of DAOD.

Our findings show that modern object detection models frequently rely on foreground–background (FG–BG) associations that often do not generalize well across domains. Through systematic background masking and feature-level perturbations, we demonstrate that removing or altering background information can lead to substantial drops in detection performance even when the foreground remains intact. These effects are consistent across various model architectures and domain pairs, including ALDI++, a state-of-the-art DAOD model on Cityscapes dataset.

Furthermore, we present that FG–BG associations are not only empirically observable but also causally linked to detection outcomes. Using a combination of do-calculus, Smooth-GradCAM++, and layer-wise feature analysis, we construct and validate a causal model that quantifies the influence of context bias. We introduce a novel domain association gradient metric and find that domain shifts exacerbate performance disparities when models rely on FG-BG associations.

\subsection*{Limitations}
Despite the strength of our analysis, we acknowledge that extracting foreground and background features separately across large datasets is computationally expensive. This limits the scalability of some of the proposed methods. Additionally, our study does not explore transformer-based architectures, which may inherently reduce FG–BG dependency due to their global receptive fields; however, interpreting FG–BG associations in such architectures remains ambiguous. Some outliers might be derived from imbalanced foreground objects of each dataset. For example, ``Car'' and ``Person'' are dominant but other foreground objects are rare. There are also certain neural network architectures such as in ``Yo'' that prevent us from computing the CAM masks and deriving the FG and BG activation features. Thus, while our method is robust, it is only applicable for architectures where we can run gradient based CAM methods.

\subsection*{Future work}
Importantly, our results suggest that current DAOD methods may unintentionally reintroduce context bias from the target domain. This highlights a new dimension of the domain adaptation problem and points to the need for bias-aware adaptation strategies that explicitly consider FG–BG association. The FG–BG association may act as a spurious or beneficial factor, depending on the stage of the pipeline. For example, during feature engineering, FG–BG bias may hinder generalization, whereas selectively leveraging it post-feature extraction could enhance performance.

We believe our work opens a novel research direction in DAOD by emphasizing the need to go beyond feature alignment and focus on understanding and mitigating causal biases introduced by background context. Future work may explore efficient integration of bias-awareness into end-to-end training pipelines and investigate connections between FG–BG associations and broader issues such as spurious correlations and fairness in DAOD.

\bibliography{tmlr}
\bibliographystyle{tmlr}

\appendix

\end{document}

%% file: math_commands.tex
\usepackage{amsmath,amsfonts,bm}

\def\eqref#1{equation~\ref{#1}}

\def\1{\bm{1}}

\DeclareMathAlphabet{\mathsfit}{\encodingdefault}{\sfdefault}{m}{sl}
\SetMathAlphabet{\mathsfit}{bold}{\encodingdefault}{\sfdefault}{bx}{n}

